# Deciding Consistency of Databases Containing Defeasible and Strict Information*

Moisés Goldszmidt
moises@cs.ucla.edu

Judea Pearl
judea@cs.ucla.edu

Cognitive Systems Lab.
Dept. of Computer Science, UCLA
Los Angeles, CA 90024

## Abstract

We propose a norm of consistency for a mixed set $X$ of defeasible and strict sentences, based on a probabilistic interpretation of these sentences. This norm establishes a clear distinction between knowledge bases depicting exceptions and those containing outright contradictions. We then define a notion of entailment based also on probabilistic considerations and provide a characterization of the relation between consistency and entailment.

We derive necessary and sufficient conditions for consistency, and provide a simple decision procedure for testing the consistency of $X$ and identifying the inconsistent subset of sentences (in the case that $X$ is inconsistent). The same procedure can also be used to test whether a sentence is entailed by $X$. Finally, it is shown that if the sentences in $X$ are *Horn* clauses, consistency and entailment can be tested in polynomial time.

## 1 Introduction

There is a sharp difference between exceptions and outright contradictions. Two statements like "typically, penguins do not fly" and "red penguins can fly", can be accepted as a description of a world in which *redness* defines an abnormal type of penguin. However, the statements "typically, birds fly" and "typically, birds do not fly" stand in outright contradiction to each other (unless birds are non existent). Whatever interpretation we give to "typically", it is hard to imagine a world containing birds in which both statements can hold simultaneously. Yet, in spite of this clear distinction, there is no formal treatment of inconsistencies in existing proposals for non-monotonic reasoning.

Consider a database $\Delta$ containing the following sentences: "all birds fly", "typically, penguins are birds" and "typically, penguins don't fly". A *circumscriptive* theory ( [McCarthy, 86]) consisting of the sentences in $\Delta$ plus the fact that Tweety is a penguin, will render the conclusion that either Tweety is a flying penguin (and therefore is an exception to the rule "typically, penguins don't fly"), or Tweety is an exception to the rule "typically, penguins are birds" and Tweety does not fly. A formalization of the database in terms of a *default* theory (see [Reiter, 80]) will render similar conclusions for our penguin Tweety. Nevertheless, the above set of rules strike our intuition as being inherently wrong: if all birds fly, there cannot be a nonempty class of objects (penguins) that are "typically birds" and yet "typically, don't fly". We cannot accept this database as merely depicting exceptions between classes of individuals; rather, it would seems that there is no possible state of affairs in which this set of sentences can hold simultaneously[1]. However, if we now change the first sentence of $\Delta$ from strict to defeasible (to read "typically, birds fly" instead of "all birds fly"), we are willing to cope with the apparent contradiction by considering the set of penguins as exceptional birds. This interpretation will remain satisfactory even if we made the second rule strict (to read "all penguins are birds"). Yet, if we further add to $\Delta$ the sentence "typically, birds are penguins" we are faced again with an intuitive *inconsistency*.

This paper deals with the problem of formalizing, detecting and isolating such inconsistencies in knowledge bases containing both defeasible and strict information[2]. We will interpret a *defeasible* sentence such as "typically, if $\phi$ then $\psi$" (written $\phi \rightarrow \psi$), as the conditional probability $P(\psi|\phi) \geq 1-\varepsilon$, where $\varepsilon > 0$ [3]. A *strict* sentence such as "if $\varphi$ it must be the case that $\sigma$" (written $\varphi \Rightarrow \sigma$), will be interpreted as the conditional probability $P(\sigma|\varphi) = 1$. Our criterion for testing inconsistency translates to that of determining if there exists a probability distribution $P$ that satisfies all these conditional probabilities for all $\varepsilon > 0$. Furthermore, to match our intuition that conditional sentences do not refer to empty classes, nor are they confirmed by merely "falsifying" their antecedents, we also require that $P$ be *proper*, i.e., that it does not render any antecedent as totally impossible. We shall show that these two requirements properly capture our intuition regarding the consistency of conditionals sentences.

---

*This work was supported in part by National Science Foundation grant #IRI-86-10155 and Naval Research Laboratory grant #N00014-89-J-2007.

[1] Provided that the set of penguins is nonempty.

[2] The consistency of systems with only defeasible sentences is analyzed in [Adams, 75] and [Pearl, 87].

We also define a notion of entailment in which plausible conclusions are guaranteed arbitrarily high probabilities in all proper probability assignments in which the defeasible premises have arbitrarily high probabilities and in which the strict premises have probabilities equal to one. A characterization of the relation between entailment and consistency is shown through the theorems of section 3.

The paper is organized as follows: section 2 introduces notation and some preliminary definitions. Consistency and entailment are explored in section 3. An effective procedure for testing consistency and entailment is presented in section 4. Section 5 contains illustrative examples, and in section 6 we summarize the main results of the paper. All proofs are given in the appendix.

## 2 Notation and Preliminary Definitions

We will use ordinary letters from the alphabet (except $d, s$ and $x$) as propositional variables. Let $\mathcal{F}$ be a language built up in the usual way from a finite set of propositional variables and the connectives "$\neg$" and "$\vee$" (the other connectives will be used as syntactic abbreviations), and let the greek letters $\phi, \psi, \varphi, \sigma$ stand for formulas of $\mathcal{F}$.

Let $\phi$ and $\psi$ be two formulas in $\mathcal{F}$. We will use a new binary connective "$\rightarrow$" to construct a defeasible sentence $\phi \rightarrow \psi$, which may be interpreted as "if $\phi$ then typically $\psi$". The set of defeasible sentences will be denoted by $D$. Similarly, given $\varphi, \sigma$ in $\mathcal{F}$, the binary connective "$\Rightarrow$" will be used to form a strict sentence $\varphi \Rightarrow \sigma$, which is to be interpreted as "if $\varphi$ then it must be the case that $\sigma$"[4]. The set of strict sentences will be denoted by $S$ [5]. We will use $X$ to stand for the union of $D$ and $S$ and $x, d, s$ as variables for sentences in $X, D$ and $S$ respectively. We will use the term *conditional* when talking about a sentence that can be either defeasible or strict. If $x$ denotes a conditional sentence with antecedent $\varphi$ and consequent $\psi$, then the *negation* of $x$, denoted by $\sim x$, is defined as a conditional with antecedent $\varphi$ and consequent $\neg\psi$. Finally, the *material counterpart* of a conditional sentence with antecedent $\phi$ and consequent $\psi$ is defined as the formula $\phi \supset \psi$ (where "$\supset$" denotes material implication).

Given a factual language $\mathcal{F}$, a *truth assignment* for $\mathcal{F}$ is a function $t$, mapping the sentences in $\mathcal{F}$ to the set $\{1, 0\}$, (1 for *True* and 0 for *False*), such that $t$ respects the usual boolean connectives [6]. A sentence $x \in X$ with antecedent $\phi$ and consequent $\psi$ will be *verified* by $t$, if $t(\phi) = t(\psi) = 1$. If $t(\phi) = 1$ but $t(\psi) = 0$, the sentence $x$ will be *falsified* by $t$. Finally, when $t(\phi) = 0$, $x$ will be considered as neither *verified* nor *falsified*.

**Definition 1** (*Probability assignment*). Let $P$ be a probability function on truths assignments, such that $\sum_j P(t_j) = 1$. We define a probability assignment $P$ on a sentence $\phi \rightarrow \psi$ from $D$ as:

$$P(\phi \rightarrow \psi) = \frac{\sum_i P(t_j) t_j(\phi \wedge \psi)}{\sum_j P(t_j) t_j(\phi)} \quad (1)$$

where $t_1, \ldots, t_i$ are all the possible truth assignments to the propositional variables in $\mathcal{F}$ and $P(t_j)$ is the probability assigned to $t_j$. We assign probabilities to the sentences in $S$ in exactly the same fashion. $P$ will be considered to be *proper*, if the denominator of Eq (1) is non-zero for every sentence in $D \cup S$.

The definition of probability assignment above, attaches a conditional probability interpretation

$$P(\psi|\phi) = \frac{P(\psi \wedge \phi)}{P(\phi)} \quad (2)$$

to the sentences in $X$. Eq. (1) states that the probability of a conditional sentence $x$ with antecedent $\phi$ and consequent $\psi$ is equal to the probability of $x$ being verified (i.e. $t_j(\phi \wedge \psi) = 1$), divided by the probability of its being either verified or falsified (i.e. $t_j(\phi) = 1$).

Up to this point the only difference between defeasible sentences and strict sentences was syntactic. They were assigned probabilities in the same fashion and they were verified and falsified under the same truth assignments. Their differences will become clear in the next section, and it rests upon the way they enter the definition of *consistency*.

---

[3] Intuitively we would like defeasible sentences to be interpreted as conditional probabilities with very high likelihood and $\varepsilon$ to be an infinitesimal quantity. For more on probabilistic semantics for default reasoning the reader is referred to [Pearl, 88].

[4] In the domain of non-monotonic multiple inheritance networks, the interpretation for the defeasible sentence $\phi \rightarrow \psi$ would be "typically $\phi$'s are $\psi$'s", while the interpretation for the strict sentence $\varphi \Rightarrow \sigma$ would be "all $\varphi$'s are $\sigma$'s".

[5] Note that both "$\rightarrow$" and "$\Rightarrow$" can occur only as the main connective.

[6] Note that if there are $n$ propositional variables in $\mathcal{F}$, there will be $2^n$ different truth assignments for $\mathcal{F}$.

# 3 Probabilistic Consistency and Entailment

In all theorems and definitions below, we will consider that the language $\mathcal{F}$ is fixed, and $d'$, $s'$, $x'$ will stand for new defeasible, strict and conditional sentences respectively, with antecedents and consequents in $\mathcal{F}$.

**Definition 2** (*Probabilistic consistency*) Let $D$ and $S$ be sets of defeasible and strict sentences respectively, constructed from formulas in $\mathcal{F}$. We say that $X = D \cup S$ is *probabilistically consistent* (p-consistent) if, for every $\varepsilon > 0$, there is a proper probability assignment $P$ such that $P(d) \geq 1 - \varepsilon$ for all defeasible sentences $d$ in $D$, and $P(s) = 1$ for all strict sentences $s$ in $S$.

Intuitively, consistency means that it is possible for all defeasible sentences to be as close to absolute certainty as desired, while the probability assignment for strict sentences is fixed at one (i.e., we have absolute certainty about the strict sentences). Another way of formulating consistency is as follows: consider a constant $\varepsilon > 0$ and let $\mathcal{P}_{X,\varepsilon}$ stand for the set of probability distributions proper for $X$ such that if $P \in \mathcal{P}_{X,\varepsilon}$ then $P(d) \geq 1 - \varepsilon$ and $P(s) = 1$ for all defeasible sentences $d \in D$ and all strict sentences $s \in S$. Consistency guarantees that $\mathcal{P}_{X,\varepsilon}$ is non-empty for every $\varepsilon > 0$.

Before developing a syntactical test for consistency (Theorem 1), we need to define the concepts of *tolerance* and *confirmation*.

**Definition 3** (*Tolerance*) Let $x$ be a sentence in $X$ with antecedent $\phi$ and consequent $\psi$. We say that $x$ is *tolerated* by the rest of the sentences in $X$, if there exists a truth assignment $t$ such that the formula $\phi \wedge \psi \wedge X_M$ is satisfied by $t$ where $X_M$ denotes the conjunction of the material counterparts of the sentences in $X$.

Thus, $x$ is tolerated by a set of conditional sentences $X$, if there is a truth assignment $t$ such that $x$ is *verified* while no sentence in $X$ is *falsified* by $t$.

**Definition 4** (*Confirmation*) We will say that a non-empty set of sentences $X = D \cup S$ is *confirmable* when:

1. If $D$ is non-empty, at least one sentence $d \in D$ is tolerated by the rest of the sentences in $X$.
2. If $D$ is empty, each sentence $s$ in $S$ is tolerated by the rest of the sentences in $S$.

**Theorem 1** Let $X = D \cup S$ be a non-empty set of defeasible and strict sentences. $X$ is p-consistent if and only if every non-empty subset of $X$ is confirmable.

Theorem 1 yields a simple decision procedure for determining p-consistency and identifying the inconsistent set in $X$ (see section 4).

Before turning our attention to issues of entailing new conditional sentences from a consistent database, we need to make explicit a particular form of inconsistency:

**Definition 5** (*Substantive inconsistency*) Let $X$ be a p-consistent set of conditional sentences, and let $x'$ be a conditional sentence with antecedent $\phi$. We will say that $X \cup \{x'\}$ is *substantively inconsistent* if $X \cup \{True \rightarrow \phi\}$ is p-consistent but $X \cup \{x'\}$ is p-inconsistent.

Non-substantive inconsistency occurs whenever the antecedent of a conditional sentence has probability equal to zero in all the probabilistic models supporting the sentences in a consistent set $X$. It will become apparent from the theorems to follow, that a set $X \cup \{x\}$ is non-substantively inconsistent iff both $X \cup \{x\}$ and $X \cup \{\sim x\}$ are inconsistent.

The concept of *entailment* introduced below is based on the same probabilistic interpretation for defeasible and strict sentences used in the definition of p-consistency and on the requirements of properness for their probabilistic models. Intuitively, we want p-entailed conclusions to receive arbitrarily high probability in all proper probability distributions in which the defeasible premises have also arbitrarily high probability, and in which the strict premises have probability equal to one.

**Definition 6** (*p-entailment*). Given a p-consistent set $X$ of conditional sentences, $X$ p-entails $d'$ (written $X \models_p d'$) if:

1. There exists a non-empty set of probability distributions which are proper for $X \cup \{d'\}$ and
2. For all $\varepsilon > 0$ there exists $\delta > 0$ such that for all probability assignments $P \in \mathcal{P}_{X,\delta}$ which are proper for $d'$, $P(d') \geq 1 - \varepsilon$.

Theorem 2 relates the notions of entailment and consistency:

**Theorem 2** If $X$ is p-consistent, $X$ p-entails $d'$ if and only if $X \cup \{\sim d'\}$ is substantively inconsistent.

Definition 6 and Theorem 3 below characterize the conditions under which conclusions are guaranteed not only very high likelihood but absolute certainty. We call this form of entailment *strict p-entailment*:

**Definition 7** (*Strict p-entailment*). If $X$ is p-consistent, then $X$ strictly p-entails $s'$ (written $X \models_s s'$) if:

1. There exists a non-empty set of probability distributions which are proper for $X \cup \{s'\}$ and
2. For all $\varepsilon > 0$, every probability assignment $P \in \mathcal{P}_{X,\varepsilon}$ that is proper for $X$ and $s'$ satisfies $P(s') = 1$.

**Theorem 3** If $X = D \cup S$ is p-consistent, $X$ strictly p-entails $\phi \Rightarrow \psi$ if and only if there exists a subset $S'$ of $S$ such that $S \cup \{True \rightarrow \phi\}$ is p-consistent and $\phi \Rightarrow \neg\psi$ is not tolerated by $S'$.

Note that strict p-entailment subsumes p-entailment, i.e., if a conditional sentence is strictly p-entailed then it is also p-entailed. Also, to test whether a conditional sentence is strictly p-entailed we need to check its status only with respect to the strict set in $X$. This confirms the intuition that we can not deduce "hard" rules from "soft" ones. However, strict p-entailment is different than logical entailment because the requirements of substantive consistency and properness for the probability distributions distinguishes strict sentences from their material counterpart. For example, consider the database $X = S = \{True \Rightarrow \neg a\}$ which is clearly p-consistent. While $X$ logically entails $a \supset b$, $X$ does not strictly p-entails $a \Rightarrow b$, since the antecedent $a$ is always falsified.

For completeness, we now present two more theorems relating consistency and entailment. Similar versions of these theorems, for the case of purely defeasible sentences, first appeared in [Adams, 75]. They follow from previous theorems and definitions.

**Theorem 4** If $X$ does not p-entail $d'$ and $X \cup \{d'\}$ is substantively inconsistent, then for all $\varepsilon > 0$ there exists a probability assignment $P \in \mathcal{P}_{X,\varepsilon}$ which is proper for $X$ and $d'$ such that $P(d') \leq \varepsilon$.

**Theorem 5** If $X = D \cup S$ is p-consistent, then

- It cannot be the case that both $d'$ and $\sim d'$ are substantively inconsistent with respect to $X$.
- It cannot be the case that both $s'$ and $\sim s'$ are substantively inconsistent with respect to any subset of $S$.

## 4 An Effective Procedure for Testing Consistency

A procedure to test the consistency of a database $X = D \cup S$ in accordance with Theorem 1 will consist of two phases: In the first phase, until $D$ is empty, we repeatedly remove a sentence from $D$ that is tolerated by the rest of the sentences in $D \cup S$. In the second phase we must test whether every sentence in $S$ is tolerated by the rest of the sentences in $S$ (without removing any sentence). If both phases can be successfully completed X is consistent, else $X$ is inconsistent.

The same procedure can be used for entailment, since to determine whether a defeasible sentence $d'$ is entailed by $X$ we need only test the consistency of $X \cup \{\sim d'\}$ and $X \cup \{d'\}$ (to make sure that the former is substantively inconsistent). The following theorem and the correctness of the procedure outlined above are proven in the appendix.

**Theorem 6** The worst case complexity of testing consistency (or entailment) is bounded by $[\mathcal{PS} \times (\frac{|D|^2}{2} + |S|)]$ where $|D|$ and $|S|$ are the number of defeasible and strict sentences respectively, and $\mathcal{PS}$ is the complexity of propositional satisfiability for the material counterpart of the sentences in the database.

Although the general satisfiability problem is NP-complete, if the sentences in $X$ are restricted to be Horn clauses then $\mathcal{PS} = O(N)$, where $N$ is the total number of occurrences of literals in $X$ [Dowling et. al., 84]. Thus, for the case of Horn clauses, testing consistency is polynomial.

## 5 Examples

**Example 1** On *birds* and *penguins*.
  We begin by testing the consistency[7] of the database presented in the introduction:

1. $b \Rightarrow f$ ("all birds fly").
2. $p \rightarrow b$ ("typically, penguins are birds")
3. $p \rightarrow \neg f$ ("typically, penguins don't fly")

Clearly none of the defeasible sentences in the example can be tolerated by the rest. If for example $t(p) = t(b) = 1$ (testing whether sentence (2) is tolerated), the assignment $t(f) = 1$ will falsify sentence (3), while the assignment $t(f) = 0$ will falsify sentence (1). A similar situation arises when we check if sentence (3) can be tolerated. Note that changing sentence (1) to be defeasible, renders the database consistent: $b \rightarrow f$ is tolerated by sentences (2) and (3) through the truth assignment $t(b) = t(f) = 1$ and $t(p) = 0$, while the remaining sentences *tolerate* each other. If we further add to this modified database the sentence $p \wedge b \rightarrow f$, we get an inconsistent set, thus showing (by Theorem 2) that $p \wedge b \rightarrow \neg f$ is p-entailed, as expected ("typically penguins_birds don't fly"). The set will become inconsistent again by adding the sentence $b \rightarrow p$ ("typically, birds are penguins"), in conformity to the graphical criteria of [Pearl, 87].

**Example 2** On *quakers* and *republicans*.
  Consider the following set of sentences:

1. $n \rightarrow r$ ("typically, Nixonites[8] are republicans")
2. $n \rightarrow q$ ("typically, Nixonites are quakers")
3. $q \Rightarrow p$ ("all quakers are pacifists")
4. $r \Rightarrow \neg p$ ("all republicans are non-pacifists")
5. $p \rightarrow c$ ("typically, pacifists are persecuted")

Sentence (5) is tolerated by all others, but the remaining sentences (1)-(4) are not confirmable. Thus this set of sentences is inconsistent. Note that Theorem 1 and the procedure outlined in the previous section not only provide a criteria to decide whether a database of defeasible and strict information is inconsistent, but also identify the *offending* set of sentences.

We can modify the above set of sentences to be:

---

[7] The terms *consistency* and *p-consistency* will be used interchangeably.

[8] "Nixonites" are members of R. Nixon's family.

137

1. $n \Rightarrow r$ ("all Nixonites are republicans")
2. $n \Rightarrow q$ ("all Nixonites are quakers")
3. $q \to p$ ("typically, quakers are pacifists")
4. $r \to \neg p$ ("typically, republicans are non-pacifists")
5. $p \to c$ ("typically, pacifists are persecuted)

This database is consistent. There is an important difference between the former case and this one. If all quakers are pacifists and all republicans are non-pacifists, our intuition immediately reacts against the idea of finding an individual that is both a quaker and a republican. On the other hand, this last set of sentences allows a "Nixonite" that is both a quaker and a republican to be either pacifist or non pacifist. Note that both $w \to p$ and $w \to \neg p$ are consistent so neither is p-entailed, and we can assert that the conclusion is ambiguous (i.e., we cannot decide whether a "Nixonite" is typically a "pacifist" or not).

Finally, if we make (2) and (4) be the only strict rules, we get a database similar in *structure* to the example depicted by network $\Gamma_6$ in [Horty et. al., 88]:

1. $n \to r$ ("typically Nixonites are republicans")
2. $n \Rightarrow q$ ("all Nixonites are quakers")
3. $q \to p$ ("typically quakers are pacifists")
4. $r \Rightarrow \neg p$ ("all republicans are non-pacifists")
5. $p \to c$ ("typically pacifists are persecuted)

Not surprisingly, the criterion of Theorem 1 renders this database consistent and $n \to \neg p$ is p-entailed in conformity with the intuition expressed in [Horty et. al., 88].

## 6 Conclusions

The probabilistic interpretation of conditional sentences yields a consistency criterion in line with human intuition. The criterion also identifies the smallest group of sentences that produces the inconsistency. A tight relation between entailment and consistency was established and an effective procedure for testing both consistency and entailment was devised.

Although our definition of p-entailment yields a rather conservative set of conclusions (e.g., one that does not permit chaining or contraposition [Pearl, 88]), it constitutes a core of plausible consequences that should be common to every reasonable system of defeasible reasoning [Pearl, 89]. For example, the notion of p-entailment was shown to be equivalent to that of entailment in preferential models semantics, whenever the sentences in $X$ are purely defeasible (see [Lehmann et. al., 88]). Consequently, the decision procedure for both p-entailment and preferential entailment should be identical (whenever $S$ is empty). It is still interesting to compare our results and procedures with those obtained from preferential models semantics for databases containing a mixture of defeasible and strict sentences.

Future work includes a graphical decision criterion for consistency in mixed inheritance networks (extending that of [Pearl, 87]), and an exploration into more powerful notions of entailment as suggested in [Pearl, 88].

## Acknowledgments

Many of the proofs, techniques and notation are extensions of those presented in [Adams, 75].

## References

[Adams, 75] Adams, E., *The Logic of Conditionals*, chapter II, Dordrecht, Netherlands: D. Reidel.

[Adams, 66] Adams, E., *Probability and The Logic of Conditionals*, in "Aspects of Inductive Logic", ed. J. Hintikka and P. Suppes, Amsterdam: North Holland.

[Dowling et. al., 84] Dowling, W. and J. Gallier, *Linear-Time Algorithms for Testing the Satisfiability of Propositional Horn Formulae*, Journal of Logic Programming, 3:267–284, 1984.

[Horty et. al., 88] Horty, J. F. and R. H. Thomason, *Mixing Strict and Defeasible Inheritance*, in Proceedings of AAAI-88, St. Paul, Minnesota.

[Lehmann et. al., 88] Lehmann, D. and M. Magidor, *Rational Logics and their Models: A Study in Cumulative Logics*, TR-8816 Dept. of Computer Science, Hebrew Univ., Jerusalem, Israel.

[McCarthy, 86] McCarthy, J., *Applications of Circumscription to Formalizing Common-sense Knowledge*, Artificial Intelligence, 13:27–39.

[Pearl, 89] Pearl, J., *Probabilistic Semantics for Nonmonotonic Reasoning: A Survey*, in Proceedings of the First Intl. Conf. on Principles of Knowledge Representation and Reasoning, Toronto, Canada, May 1989, pp. 505–516.

[Pearl, 88] Pearl, J., *Probabilistic Reasoning in Intelligent Systems: Networks of Plausible Inference*, chapter 10, Morgan Kaufmann Publishers Inc.

[Pearl, 87] Pearl, J., *Deciding Consistency in Inheritance Networks*, Tech. Rep. (R-96) Cognitive Systems Lab., UCLA.

[Reiter, 80] Reiter, R., *A Logic for Default Reasoning*, Artificial Intelligence, 13:81–132.

# A  Appendix: Theorems and Proofs

**Theorem 1** Let $X = D \cup S$ be a non-empty set of defeasible and strict sentences constructed from the formulas in $\mathcal{F}$. $X$ is p-consistent if and only if every non-empty subset of $X$ is confirmable.

**Proof of the *only if* part:** We want to show that if there exists a non-empty subset of $X$ which is not confirmable, then $X$ is not p-consistent. The proof is facilitated by introducing the notion of *quasi-conjunction* ( [Adams, 75]): Given a set of defaults $D = \{\phi_1 \rightarrow \psi_1, \ldots, \phi_n \rightarrow \psi_n\}$ the *quasi-conjunction* of $D$ is the defeasible sentence,

$$C(D) = [\phi_1 \vee \ldots \vee \phi_n] \rightarrow [(\phi_1 \supset \psi_1) \wedge \ldots \wedge (\phi_n \supset \psi_n)] \tag{3}$$

The quasi-conjunction $C(D)$ bears interesting relations to the set $D$. In particular, if $D$ is confirmed by some assignment $t$, $C(D)$ will be verified by $t$. This is so because the verification of at least one sentence of $D$ by $t$ guarantees that the antecedent of $C(D)$ (i.e. the formula $[\phi_1 \vee \ldots \vee \phi_n]$ in Eq. (3)) is mapped into 1, and the fact that no sentence in $D$ is falsified guarantees that the consequent of $C(D)$ (i.e. the formula $[(\phi_1 \supset \psi_1) \wedge \ldots \wedge (\phi_n \supset \psi_n)]$ in Eq. (3)) is also mapped into 1. Similarly, if at least one sentence of $D$ is falsified, its quasi-conjunction is also falsified. In this case, the consequent of $C(D)$ is mapped into 0 since at least one of the material implication in the conjunction is falsified. Additionally, let $U_p(C(D)) = 1 - P(C(D))$ (the *uncertainty* of $C(D)$) where $P(C(D))$ is the probability assigned to the quasi-conjunction of $D$ according to Eq. (1), then, it is shown in [Adams, 66] that the uncertainty of the quasi-conjunction of $D$ is less or equal to the sum of the uncertainties of each of the sentences in $D$, i.e. $U_p(C(D)) \leq \sum_i (1 - P(d_i))$ where the sum is taken over all $d_i$ in $D$.

We are now ready to proceed with the proof. Let $X' = D' \cup S'$ be a subset of $X$ where $D'$ is a subset of $D$ and $S'$ is a subset of $S$. If $X'$ is not confirmable then one of the following cases must occur:

Case 1.- $S'$ is empty and $D'$ is not confirmable[9]. In this case, the quasi-conjunction for $D'$ is not verifiable; from Eq. (1), we have that $P(C(D')) = 0$ and $U_p(C(D')) = 1$. It follows, by the properties of the quasi-conjunction outlined above that $\sum_i (1 - P(d_i'))$ over all $d_i'$ in $D'$ is at least 1. If the number of sentences in $D'$ is $n > 1$, then,

$$n - \sum_{i=1}^{n} P(d_i') \geq 1 \tag{4}$$

$$\sum_{i=1}^{n} P(d_i') \leq n - 1 \tag{5}$$

which implies that at least one sentence in $D'$ has probability smaller than $1 - \frac{1}{n}$. Hence, it is impossible to have $P(d') \geq 1 - \varepsilon$, for every $\varepsilon > 0$, for every defeasible sentence $d' \in D'$. Thus, $X$ is p-inconsistent.

Case 2.- $D'$ is empty. Proof by contradiction: assume that $S'$ is not confirmable and $X'$ is p-consistent. If $X'$ is p-consistent, there must exist a probability assignment $P$ satisfying definition 2, and a set $T$ of truth assignments such that $P(t_i) > 0$ for all $t_i$ in $T$. If $S'$ is not confirmable, then either one of the following conditions must be true: there is at least one truth assignment $t'$ in $T$ such that $t'$ falsifies a sentence $s'$ in $S'$, or there is a sentence $s''$ in $S'$ such that no truth assignment $t''$ in $T$ verifies $s''$. The requirements of p-consistency state that for every sentence $\varphi \Rightarrow \sigma$ in $S$, $P(\varphi \Rightarrow \sigma) = 1$. Thus, from Eq. (1),

$$P(\varphi \Rightarrow \sigma) = \frac{P(t_1)t_1(\varphi \wedge \sigma) + \cdots + P(t_n)t_n(\varphi \wedge \sigma)}{P(t_1)t_1(\varphi) + \cdots + P(t_n)t_n(\varphi)} = 1 \tag{6}$$

which immediately implies that, no sentence $s' \in S'$ can be falsified by any $t \in T$. Hence, the first condition for the unconfirmability of $S'$ cannot occur. On the other hand, if there is no $t''$ in $T$ that verifies (nor falsifies) a sentence $s''$ in $S'$, the denominator of $P(s'')$ is 0 (see Eq.( 1)), and $P$ is not proper as required. Since by the definition of confirmability these two are the only conditions under which a set of purely strict sentences can be unconfirmable, we conclude that $S'$ cannot be confirmable while $X$ is p-consistent.

Case 3.- Neither $D'$ nor $S'$ are empty and $X'$ is not confirmable. That is, either $D'$ is not confirmable or every $t'$ in $T$ that verifies a sentence in $D'$ falsifies at least one sentence in $S'$. The first situation will lead us back to case 1 while the second to a contradiction similar to case 2 above. In either case, $X$ is not p-consistent.

**Proof of the *if* part:** Assume that every non-empty subset of $X = D \cup S$ is confirmable. Then the following two constructions are feasible:

---

[9] This case is covered by Theorem 1.1 in [Adams, 75].

- We can construct a finite "nested decreasing sequence" of non-empty subsets of X, namely $X_1, \ldots, X_m$, ($X = X_1$), and an associated sequence of truth assignments $t_1, \ldots, t_m$ confirming $X_1, \ldots, X_m$ respectively, with the following characteristics:
    1. $X_{i+1}$ is the proper subset of $X_i$ consisting of all the sentences of $D_i$ not verified by $t_i$, for $i = 1, \ldots, m-1$.
    2. All sentences in $D_m$ are verified by $t_m$.
- We can construct a sequence $t_{m+1}, \ldots, t_n$ that will confirm $X_{m+1} = S$. That is, the sequence $t_{m+1}, \ldots, t_n$ will verify every sentence in $S$ without falsifying any. We will associate with $t_{m+1}, \ldots, t_n$ the "nested decreasing sequence" $X_{m+1}, \ldots, X_n$ where $X_{i+1}$ is the proper subset of $X_i$ consisting of all the sentences of $S_i$ not verified by $t_i$ for $i = m+1, \ldots, n$.

We can now assign probabilities to the truth-assignments $t_1, \ldots, t_n$ in the following way:
For $i = 1, \ldots, n-1$

$$P(t_i) = \varepsilon^{i-1}(1-\varepsilon) \tag{7}$$

and

$$P(t_n) = \varepsilon^{n-1} \tag{8}$$

We must show that, in fact, every sentence $d$ in $D$ obtains $P(d) \geq 1 - \varepsilon$ and that every sentence $s$ in $S$ obtains $P(s) = 1$. Since every sentence $d$ is verified in at least one of the member of the sequence $X_1, \ldots, X_n$, using Eq. (1) we have that for $i < n$:

$$P(d_i) \geq \frac{\varepsilon^{i-1}(1-\varepsilon)}{\varepsilon^{i-1}(1-\varepsilon) + \varepsilon^i(1-\varepsilon) + \cdots + \varepsilon^{n-1}} = 1 - \varepsilon \tag{9}$$

and $P(d_n) = 1$ if it is only verified by the last truth assignment when $S$ is originally empty. Finally, since no sentence $s$ in $S$ is ever falsified by the sequence of truth assignments $t_1, \ldots, t_n$ and each and every $s$ in $S$ is verified at least once, it follows from Eq (1) and the process by which we assigned probabilities to $t_1, \ldots, t_n$ that indeed $P(s) = 1$ for every $s \in S$.

**Theorem 2** If $X$ is p-consistent, $X$ p-entails $d'$ if and only if $X \cup \{\sim d'\}$ is substantively inconsistent.

**Proof of the *only if* part:** (If $X$ p-entails $d'$ then $X \cup \{\sim d'\}$ is substantively inconsistent.) Let $X \models_p d'$. ¿From the definition of p-entailment, if $X \models_p d'$ then for all $\varepsilon > 0$ there exists a $\delta > 0$ such that for all $P \in \mathcal{P}_{X,\delta}$ which are proper for $X$ and $d'$, $P(\sim d') \leq \varepsilon$. This means that for all proper probability assignments $P$ for $X$ and $d'$ [10], the sentence $\sim d'$ gets an arbitrarily low probability whenever all defeasible sentences in $X$ can be assigned arbitrarily high probability and all strict sentences in $X$ can be assigned probability equal to 1. Thus $X \cup \{\sim d'\}$ is substantively inconsistent.

**Proof of the *if* part:** (If $X \cup \{\sim d'\}$ is substantively inconsistent then $X$ p-entails $d'$.) Let $X \cup \{\sim d'\}$ be substantively inconsistent. From Theorem 1, we know that there must be a subset $X'$ of $X \cup \{\sim d'\}$ that is not confirmable. Furthermore, since $X$ is p-consistent, $X' = X'' \cup \{\sim d'\}$. Let $\mathcal{P}_S$ stand for the set of probability distributions that are proper for $X$ and $\sim d'$ such that if $P \in \mathcal{P}_S$, then $P(s) = 1$ for all $s$ in $X$ [11]. We will consider two cases depending on the structure of $X''$:

Case 1.- $X''$ does not include any defeasible sentences. ¿From Theorem 1, we know that $\sim d'$ cannot be tolerated by $X''$ for otherwise $X'$ wouldn't be inconsistent. It follows from Eq. (1) (probability assignment) that $P(\sim d') = 0$ for all $P \in \mathcal{P}_S$. Thus, $P(d') = 1$ in all $P \in \mathcal{P}_S$ and since any probability distribution that is in $\mathcal{P}_{X,\varepsilon}$ must also belong to $\mathcal{P}_S$, it follows from the definition of p-entailment that $X \models_p d'$.

Case 2.- $X''$ includes defeasible and a possible empty set of strict sentences. Since $X'' \cup \{\sim d'\}$ is substantively inconsistent, from the proof of Theorem 1, the following must be true for all probability distributions $P \in \mathcal{P}_S$:

$$\sum_{d \in X''} U_p(d) + U_p(\sim d') \geq 1 \tag{10}$$

which implies that

$$\sum_{d \in X} U_p(d) \geq 1 - U_p(\sim d') = U_p(d') \tag{11}$$

---

[10] Note that from the definition of *p-entailment* there must exists at least one $P$ proper for $X$ and $d'$.

[11] We know that $\mathcal{P}_S$ is not empty due to the first condition of substantive inconsistency, as applied to $X \cup \{\sim d'\}$. Also in the case where $X$ does not contain any strict sentences, $\mathcal{P}_S$ simply denotes all probability distributions that are proper for $X$ and $\sim d'$.

Since $U_p(d) = 1 - P(d)$ and $U_p(d') = 1 - P(d')$, Eq. (11) says that $1 - P(d')$ can be made arbitrarily small by requiring the values $1 - P(d)$ for $d \in D$ to be sufficiently small and the values of $P(s)$ to be 1 for all $s \in S$. This is equivalent to say that $X \models_p d'$.

**Theorem 3** If $X = D \cup S$ is p-consistent, $X$ strictly p-entails $\phi \Rightarrow \psi$ if and only if there exists a subset $S'$ of $S$ such that $S' \cup \{True \rightarrow \phi\}$ is p-consistent and $\phi \Rightarrow \neg\psi$ is not tolerated by $S'$.

**Proof** It follows from the proof of Theorem 2 (see case 1 of the *if* part).

**Theorem 6** Let $\mathcal{F}$ be a set of propositional formulas, and let $X = D \cup S$ be the a set of defeasible and strict sentences constructed from the formulas in $\mathcal{F}$. The worst case complexity of testing the consistency of $X$ is bounded by $[\mathcal{PS} \times (\frac{|D|^2}{2} + |S|)]$ where $|D|$ and $|S|$ are the number of defeasible and strict sentences respectively, and $\mathcal{PS}$ is the complexity of propositional satisfiability for the material counterpart of the sentences in $X$.

**Proof** The following procedure for testing consistency finds a "nested decreasing sequence", (see proof of Theorem 1), if one exists; otherwise, it returns failure.

```
PROCEDURE TEST_CONSISTENCY
INPUT: a set X = D ∪ S of
       defeasible and strict sentences
1.  LET D' := D
2.  WHILE D' is not empty DO
3.    Find a sentence d ∈ D' such that
      d is tolerated by S ∪ D'
4.    IF d is found then
            LET D' := D' - d
      ELSE HALT: the set is
            INCONSISTENT
ENDWHILE
5.  LET S' := S
6.  WHILE S' is not empty DO
7.    Pick any sentence s ∈ S' and test
      if s is tolerated by S
8.    IF s is tolerated then
            LET S' := S' - s
9.    ELSE HALT: the set is
            INCONSISTENT
ENDWHILE
10. The set is CONSISTENT
END PROCEDURE
```

If the procedure stops at either line (4) or line (9) a non confirmable subset is found, and by theorem 1 the set of sentences is inconsistent. On the other hand, if the procedure reaches line (10), $X$ cannot possibly contain a subset that is not confirmable. Any such subset (see Definition 4) would have halted the procedure either at line (4) or at line (9), thus, by Theorem 1, $X$ must be consistent. It follows that the procedure is correct.

To assess the time complexity, note that the WHILE-loop of line (6) will be executed $|S|$ times in the worst case, and each time we must do at most $\mathcal{PS}$ work to test the satisfiability of $S - s$; thus, its complexity is $|S| \times \mathcal{PS}$. In order to *find* a tolerated sentence $d = \phi \rightarrow \psi$ in $D'$, we must test at most $|D'|$ times (once for each sentence $d \in D'$) for the satisfiability of the conjunction of $\phi \wedge \psi$ and the material counterparts of the sentences in $S \cup D' - \{d\}$. However, the size of $D'$ is decremented by at least one sentence in each iteration of the WHILE-loop in line (2), therefore the number of times that we test for satisfiability is $|D| + |D| - 1 + |D| - 2 + \ldots + 1$ which is bounded by $\frac{|D|^2}{2}$. Thus, the overall time complexity is $O[\mathcal{PS} \times (\frac{|D|^2}{2} + |S|)]$.